\RequirePackage{scrlfile}
\PreventPackageFromLoading{subfig}
\documentclass[letterpaper, 10 pt, conference]{ieeeconf}  

\IEEEoverridecommandlockouts                   
\usepackage[table,xcdraw]{xcolor}
\usepackage{siunitx}
\usepackage{graphics} 
\usepackage{epsfig} 
\usepackage{float}
\usepackage{mathptmx} 
\usepackage{times} 
\usepackage{amsmath} 
\usepackage{amssymb}  
\usepackage{subcaption}
\usepackage{svg}
\usepackage{hyperref}
\usepackage{pgf,tikz}
\usepackage{mathrsfs}
\usetikzlibrary{arrows}
\usepackage{xcolor}
\usepackage{stackengine,scalerel}
\usepackage[pdf]{pstricks}
\usepackage[table,xcdraw]{xcolor}
\usepackage{commath}
\usepackage{subfig}
\usepackage{multirow}
\usepackage[table,xcdraw]{xcolor}
\usepackage{gensymb}

\usepackage[T1]{fontenc}
\usepackage[utf8]{inputenc}
\usepackage[none]{hyphenat}
\stackMath

\title{\LARGE \bf Deep Sensor Fusion for Real-Time Odometry Estimation 
}
 
\author{Michelle Valente$^{1}$, Cyril Joly$^{1}$ and Arnaud de La Fortelle$^{1}$% <-this % stops a space
\thanks{*This work was supported by the International \emph{Chair Drive for All} with sponsors PSA Peugeot Citroën - Safran - Valeo}% <-this % stops a space
\thanks{$^{1}$Michelle Valente, Cyril Joly and Arnaud de la Fortelle are with Center for Robotics, MINES ParisTech, PSL Research University, 60 boulevard Saint Michel, 75006 Paris, France. {\tt\small \{michelle.valente, cyril.joly, arnaud.de\_la\_fortelle\}@mines-paristech.fr}}%
}%

\begin{document}
% \twocolumn[{%
% \renewcommand\twocolumn[1][]{#1}%
\maketitle
% \begin{center}
%     \centering
%     \includegraphics[height=3.5cm]{images/teaser4.png}
%     \captionof{figure}{Test caption}
% \end{center}%
% }]

%%%%%%%%%%%%%%%%%%%%%%%%%%%%%%% Abstract %%%%%%%%%%%%%%%%%%%%%%%%%%%%%%%%%%%%%
\begin{abstract}

Cameras and 2D laser scanners, in combination, are able to provide low-cost, light-weight and accurate solutions, which make their fusion well-suited for many robot navigation tasks. However, correct data fusion depends on precise calibration of the rigid body transform between the sensors. In this paper we present the first framework that makes use of Convolutional Neural Networks (CNNs) for odometry estimation fusing 2D laser scanners and mono-cameras. The use of CNNs provides the tools to not only extract the features from the two sensors, but also to fuse and match them without needing a calibration between the sensors. We transform the odometry estimation into an ordinal classification problem in order to find accurate rotation and translation values between consecutive frames. Results on a real road dataset show that the fusion network runs in real-time and is able to improve the odometry estimation of a single sensor alone by learning how to fuse two different types of data information.  

\end{abstract}
%%%%%%%%%%%%%%%%%%%%%%%%%%%%%%%%%%%%%%%%%%%%%%%%%%%%%%%%%%%%%%%%%%%%%%%%%%%%%%

%%%%%%%%%%%%%%%%%%%%%%%%%%%%%%% Introduction %%%%%%%%%%%%%%%%%%%%%%%%%%%%%%%%%
\section{Introduction}

Self-localization of an intelligent vehicle is still a challenging and ongoing task for the autonomous driving development. A reliable localization is necessary for intelligent vehicles to be able to take important decisions like overtaking another vehicle or simply defining a trajectory. To address this task, methods known as Simultaneous Localization and Mapping (SLAM) allow to localize the vehicle in a previously unknown environment while concurrently mapping the environment. Different types of sensors can be used for this purpose, such as cameras, laser scanners and radars. Each sensor has its own particular limitations and advantages and, for this reason, a single sensor cannot reliably be used alone; thus it is necessary to perform the fusion of different sensors to increase the accuracy of various tasks. One of the main difficulties to perform the fusion is to have a common data format and an accurate calibration between the different sensors.

It is very popular the usage of 3D laser scanners for autonomous driving research, however because of its cost it has become a major drawback for automobile manufacturers to be able to maintain a reasonable price for the future vehicles. Moreover, the amount of data provided by 3D laser scanners requires large computational resources making it difficult to have real-time localization methods. Considering this, we chose to focus on a solution for vehicles equipped with low cost 2D laser scanners. On the other hand, since only a 2D slice of the surrounding environment is detected at each scan, some situations can be challenging if we only rely on this sensor. To address this problem, we propose to fuse the 2D laser scanner with a mono-camera, that can detect more information about the environment and increase the accuracy of the localization method.  

In the last years, Deep Learning methods have received attention in the field of autonomous driving. The most common applications use cameras or laser scanners for tasks such as obstacle detection \cite{detection_lidar} and classification \cite{classification}. These tasks are mainly in the field of environment understanding and mapping. However, the localization of the vehicle and the fusion of different sensors is still a task that has not yet been extensively explored by machine learning techniques. The use of Neural Networks for this purpose could not only make the task faster, but also eliminate the need of a precise calibration between the sensors.

\begin{figure}
	\centering
	\includegraphics[width=0.85\linewidth]{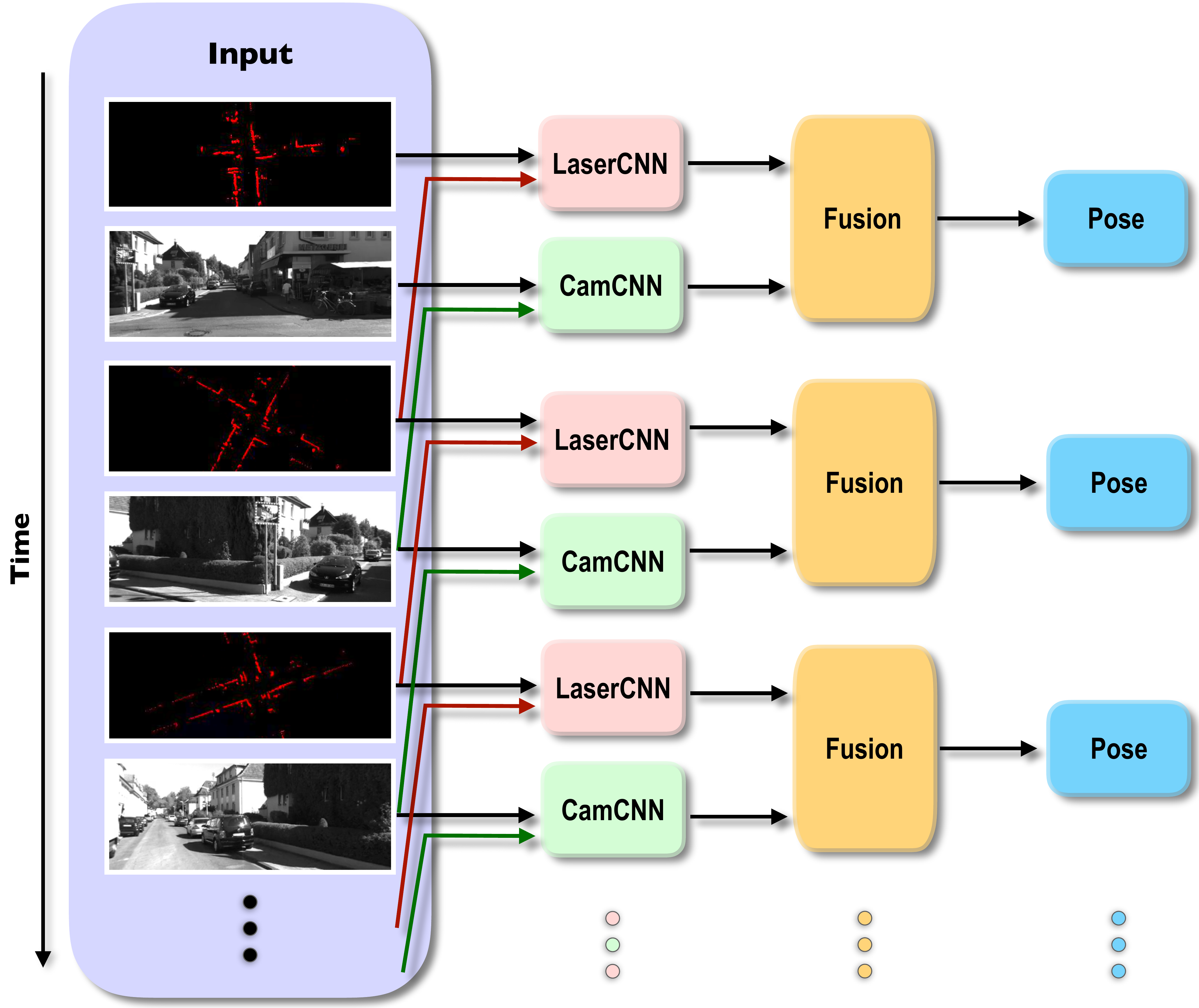}
	\caption{Overview of the proposed system. Consecutive laser scans and camera images are used, first in two separate networks and sequentially in a common network, in order to estimate the pose of the vehicle between consecutive frames. }
	\label{fig:System}
\end{figure}

Considering this, we propose a complete Deep Learning approach to use CNNs to extract the features of two different sensors, a 2D laser scanner and a mono-camera, and sequentially fuse them in order to estimate the odometry of the vehicle. A schematic of the proposed approach is presented in Figure~\ref{fig:System}. We also present a new method to transform the odometry estimation regression problem into an ordinal classification that facilitates the training of the network. The solution can be run in real time and tests in real road scenarios shows competitive results compared to other deep learning approaches. 

The remainder of the paper is organized as follows. First, we present the related work in Section~\ref{relatedwork.sec}; the proposed method and the design of the network is presented in Section~\ref{method.sec}; experimental results are presented in Section~\ref{results.sec}; finally conclusion and perspectives are given in Section~\ref{conclusion.sec}.

%%%%%%%%%%%%%%%%%%%%%%%%%%%%%%%%%%%%%%%%%%%%%%%%%%%%%%%%%%%%%%%%%%%%%%%%%%%%%%
%%%%%%%%%%%%%%%%%%%%% Related Works %%%%%%%%%%%%%%%%%%%%%%%%%%%%%%%%%%%%%%%%%%
\section{Related Work}\label{relatedwork.sec}

The application of Deep Learning techniques has presented impressive results recently in tasks such as object detection and classification by the use of camera images. To obtain these results, a large amount of datasets has become available in the last years. In the context of intelligent vehicles, interesting work has emerged in different fields: mapping \cite{mapping}, trajectory prediction \cite{trajectory}, control \cite{control} and even end-to-end approaches \cite{endtoend}.

At the same time, localization, which is still a very challenging problem for robotic systems, is not yet well explored by deep learning methods. The most common approaches are based on the use of camera images for odometry estimation. They are inspired by the classic methods for Visual Odometry (VO) 
\cite{orb}\cite{svo}, which consists in estimating the camera's motion by finding geometry constraints from a sequence of images. The use of machine learning for this purpose allows to deal with challenging environments and camera parameters difficulties. The first method proposed was PoseNet \cite{posenet}, which it is based on the use of CNNs to estimate the 6-DoF pose using only RGB images. More recently, Wang et al. \cite{deepvo} introduced the DeepVO method, which uses RCNNs with the same goal. The same authors also presented the method UndeepVO \cite{undeepvo}, which proposes an unsupervised deep learning method to estimate the pose of a monocular camera. However, the classic VO methods still outperform deep learning based methods published to this date, considering the accuracy in the pose estimation.

Laser scanners are also popular for classic pose estimation because of its accuracy. Classic approaches for this problem consist in trying to match two point clouds and estimate the transformation between them, this solution is known as Iterative Closest Point (ICP) \cite{icp}. Since then, more robust and complex algorithms were presented to achieve the same objective. LOAM \cite{loam} is currently one of the most popular method because of its high accuracy and ability to achieve real-time processing by running two different algorithms in parallel. 

The application of deep learning techniques for this purpose using laser scanners is still considered as a new challenge and only few papers have addressed it. Nicolai et al. \cite{laser1} were the first to propose to apply 3D laser scanner data in CNNs to estimate odometry. Their approach provides a reasonable estimation of odometry, however still not competitive with the efficiency of state-of-the-art scan matching methods. Later, Velas et al. \cite{laser3} presented another approach for using CNNs with 3D laser scanners for IMU assisted odometry. Their results were able to get high precision and close results compared to state-of-the-art methods, such as LOAM \cite{loam}, for translation, however the method is not able to estimate rotation with sufficient precision. Considering their results, the authors propose that their method could be used as a translation estimator and use together an Inertial Measurement Unit (IMU) to obtain the rotation. Another drawback is that, according to the KITTI benchmark, even using CNNs the method is slower than LOAM.

The use of 2D laser scanners instead of 3D laser scanners could reduce considerably the price of future intelligent vehicles and the need of high computational resources. The authors of this work presented in \cite{previouswork} a solution that relies only on a 2D laser scanner for odometry estimation using Recurrent Convolutional Neural Networks (RCNNs). The method showed promising results along with a very fast computational time. However, one of the main difficulties encountered by this approach was that sometimes in challenging environments the 2D laser scanner could not detect many obstacles and it would generate inaccurate results. Considering this, in this work we propose a new solution to improve the odometry estimation by fusing mono-cameras and 2D laser scanners using only Convolutional Neural Networks.

The fusion between laser scanners and cameras is commonly used in tasks such as object recognition and navigation of mobile robots \cite{fusion3}\cite{fusion2}. The use of Deep Learning with sensor fusion is not yet extensively explored; Most of the existing deep sensor fusion solutions are based on object detection, like in \cite{pointfusion} where the authors fuse 3D laser scanners and camera images to predict object's bounding boxes. There are also methods that use sensor fusion for end-to-end learning \cite{stearingfusion}, where the input is the data of different sensors and the output is directly steering commands. 

We present the first deep learning method for odometry estimation based on the fusion of a 2D laser scanner and a camera. The proposed network is able to provide a real-time solution that overcome the difficulties enconter by one sensor alone. We also present how to transform the odometry regression problem into a into a series of simpler binary classification subproblems, known as ordinal classification. Finally, we explore this solution in outdoor environments, training and testing it with the KITTI \cite{kitti} dataset, which contains sequences on different types of scenarios.

\begin{figure*}
	\centering
	\includegraphics[width=0.95\linewidth]{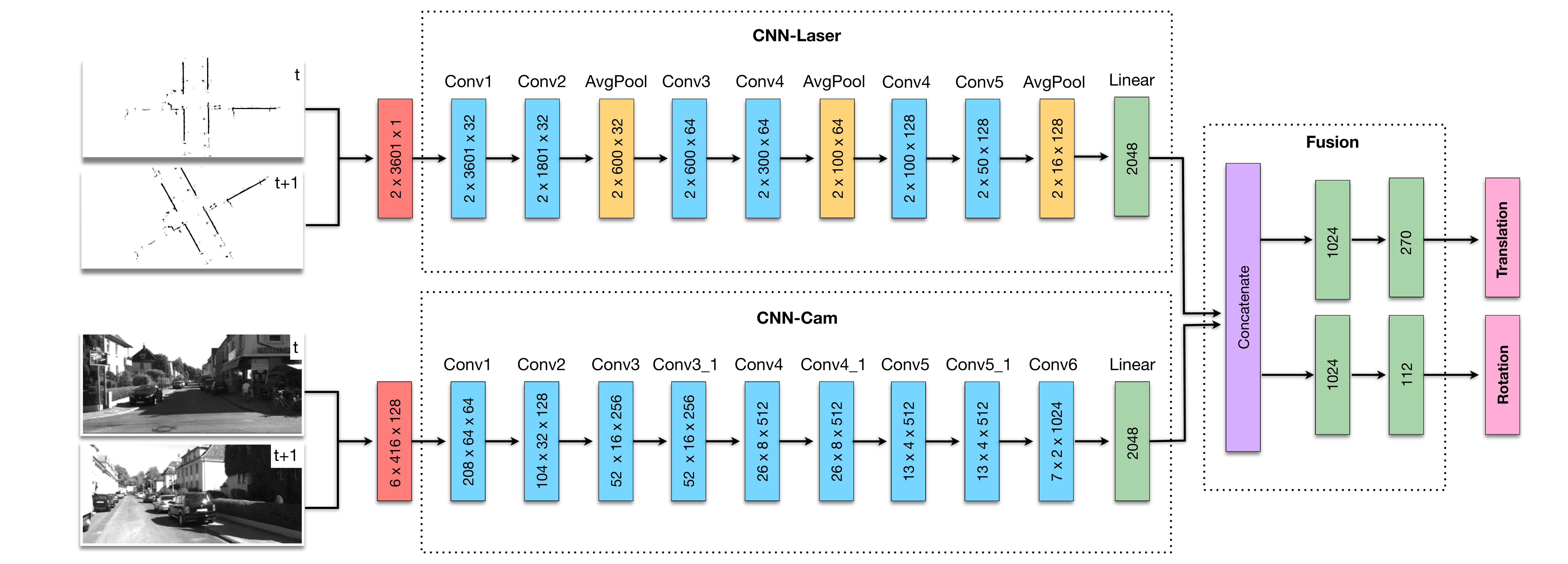}
	\caption{Architecture of the proposed network. Each block of the illustration presents the size of the tensors considering that the input is the raw sensor data pre-processed as presented in \autoref{sub.encoding}. }
	\label{fig:architecture}
\end{figure*}

%%%%%%%%%%%%%%%%%%%%%%%%%%%%%%%%%%%%%%%%%%%%%%%%%%%%%%%%%%%%%%%%%%%%%%%%%%%%%%
%%%%%%%%%%%%%%%%%%%%%%%%%%%%%%% Method %%%%%%%%%%%%%%%%%%%%%%%%%%%%%%%%%%%%%%%
\section{Method}\label{method.sec}

The proposed approach consists in finding the vehicle displacement by estimating the transformation between a sequence of camera images and laser scanner acquisitions. From two consecutive observations, where each observation is a $\ang{360}$ set of points measured during one laser rotation and one camera image, the network predicts the transformation $T = [\Delta{d}, \Delta{\theta}]$, which represents the travelling distance $\Delta{d}$ and the orientation $\Delta{\theta}$ between two consecutive laser scans $(s_{t-1}, s_{t})$ and camera images $(c_{t-1}, c_{t})$. Therefore, the goal is to learn the optimal function $g(.)$, which maps the fusion between $(s_{t-1}, s_{t})$ and $(c_{t-1}, c_{t})$ to $T$ at time $t$:
\begin{equation}
    T_t = g((s_{t-1}, s_{t}), (c_{t-1}, c_{t}))
\label{learning_function}
\end{equation}

Once we learn these parameters, we can obtain the 2D pose $(x_t,y_t,\theta_t)$ of the vehicle in time $t$  as follow:
\begin{equation}
\begin{split}
    x_t& = x_{t-1} + \Delta{d} \sin{(\Delta{\theta})} \\
    y_t& = y_{t-1} + \Delta{d} \cos{(\Delta{\theta})} \\
    \theta_t& = \theta_{t-1} + \Delta{\theta}
\end{split}
\end{equation}

In this way, we can accumulate the local poses of the vehicle and estimate the global position of the vehicle at any time $t$. Since the algorithm does not perform any sort of loop closure, drift can be also accumulated, thus reducing the accuracy of the vehicle's localization. The main goal is to explore the use of CNNs to match laser scans and camera images between two consecutive frames for odometry estimation, and especially, to prove that the fusion between the two of them for this purpose is possible to be executed by neural networks.

The following subsections will present in details the proposed method. First, we show how the raw data of the sensors are pre-processed. Sequentially, we present the configuration of the network and the specifics of the training process. 

\subsection{Data Pre-processing}\label{sub.encoding}

The raw data coming from the two sensors need to be prepared before they can be used as an input for the neural network. For the laser scanner we use the same data encoding of our previous work \cite{previouswork}, where the sensor point set is encoded into a 1D vector. The vector is created by binning the raw scans into bins of resolution $\ang{0.1}$. Since many points can fall into the same bin, we calculate the average depth value. Finally, considering all the bins of a $\ang{360}$ rotation range, we store the depth values into a 3601 size vector, where each possible bin angle average depth is represented by the elements in the vector. 

After two sequential laser scans are encoded as 1D vectors, we concatenate them to create the input of the laser scanner network. The idea is to create a sort of image of size $(2,3601)$, allowing to use standard convolutional layers to extract the features detected by the sensor in the surrounding environment. 

For the camera raw data we only resize the images in order to reduce the computational time. We tested different sizes so that the accuracy of the method was not reduced but we would still be able to produce faster results. Considering this, the best match for accuracy and time was achieved with the image size $(416, 128)$. After resizing them, two consecutive image are stacked together, in the same way of the scans, to form a tensor that represents two camera acquisitions. This format allows us to feed the two images to the same sequence of CNNs to extract the features.

\subsection{Network Architecture}

In \autoref{fig:architecture} the architecture of the proposed network is presented. The input consists in the pre-processed raw data from the two sensors as explained in the previous subsection, while the rest of the network can be separated in the three main parts explained bellow: the laser scanner (CNN-Laser), the camera (CNN-Cam) and finally the fusion. 

\subsubsection{CNN-Laser} Two pre-processed laser scanner acquisitions, represented as a one dimension vector of size 3601, are concatenated to create the input tensor of the network. Sequentially, the tensor is fed into the sequence of 1D convolutional and average pool layers to learn the features between the two acquisitions. We use the same CNN configuration of our previous work \cite{previouswork}. It consists of 6 1D convolutional layers, where each layer is followed by a rectified linear unit (ReLU) activation. Between each sequence of two convolutional layers, there is also one average pool layer to reduce computation complexity. The difference from the previous work is the extra layer, the linear layer, after the sequence of CNNs to reduce the tensor size before the fusion. The goal is to input the same amount of information from both of the sensors to the fusion part of the network. 

\subsubsection{CNN-Cam} The configuration of the camera network is the CNN part of the RCNN proposed by DeepVO \cite{deepvo}. However, we use a smaller input as explained before in the pre-processing. In the same way of the CNN-Laser, we added an extra linear layer to reduce the tensor size before the fusion. Therefore, the input is two raw camera images and the output is the reduced features detected by the sequence of CNNs to be then fused with the features detected at the CNN-Laser. 

\subsubsection{Fusion} After extracting features from consecutive laser scans and camera images using the previous described CNNs, we concatenate their outputs in order to estimate the pose of the vehicle. The concatenated features are fed to two different sequence of linear layers. We tested the use of same linear layers for both rotation and translation estimation together, however better results were found once we separated them. In order to avoid the overfitting, the two linear layers are preceded by Dropout layers.  

\subsection{Training}
\label{sub:training}

The training of the network was performed in two steps; First, we trained the single sensor CNNs separately to find the best pre-trained weights possible for those networks. For this purpose, we connected them to two separate linear layers, one for the rotation and one for the translation. After it, the pre-trained CNNs are connected to the fusion part of the network and we perform the final training step.

In \cite{previouswork}, we prove that the CNN network get better results if we reformulate the problem as a classification task for the rotation, and continue it as a regression one for the translation. Considering all the possible variation of angles between two frames, we created classes for the interval $\pm\ang{5.6}$ with $\ang{0.1}$ resolution, resulting in 112 possible classes. 

In this work we propose to not only treat the rotation as a classification task, but also the translation in order to facilitate the training. We observed that in all the possible sequences of the KITTI dataset the maximum translation between two frames is around $2.6$ meters and the minimum $0.0$. Therefore, we created 270 classes for this interval considering the resolution of $0.01$ meters. Results showed that it became easier for the network to converge and the accuracy was not reduced by this transformation.

The main problem of transforming the rotation and translation in a classification task is that no order about the data is learned by the network. This happens because machine learning methods for classification problems commonly assume that the class values are unordered. For example, it would not be possible to understand that a difference of $2$ degrees was higher than only $0.1$ degrees. In \cite{ordinal} the authors introduce a simple method that enables standard classification algorithms to make use of ordering information in class attributes, known as ordinal classification. The idea is to transform the ordinal regression problem into a series of simpler binary classification subproblems. Inspired by this work, the authors of \cite{ordinal2} applied this idea solving the simpler binary classifications for age estimation by the use of CNNs. 

In order to transform our rotation and translation classes into a series of binary classification subproblems it is necessary to change how we label the dataset. Instead of labeling directly with a class that represents the ground truth value, the samples are labeled by an ordinal scale called the rank. Considering $k$ the number of possible classes, we are going to transform the problem into $k-1$ simpler binary classification subproblems. Specifically, for one sample $i$ each subproblem receives a label $l^k_i \in {0,1}$ indicating if the sample class $l_i$ is larger than $r_k$ as follow,
\begin{equation}
    l^k_i = 
    \begin{cases}
    1, \textit{ if }(l_i > r_k) \\
    0, \textit{ otherwise.}
    \end{cases}
\end{equation}

\begin{figure}
	\centering
	\includegraphics[width=0.9\linewidth]{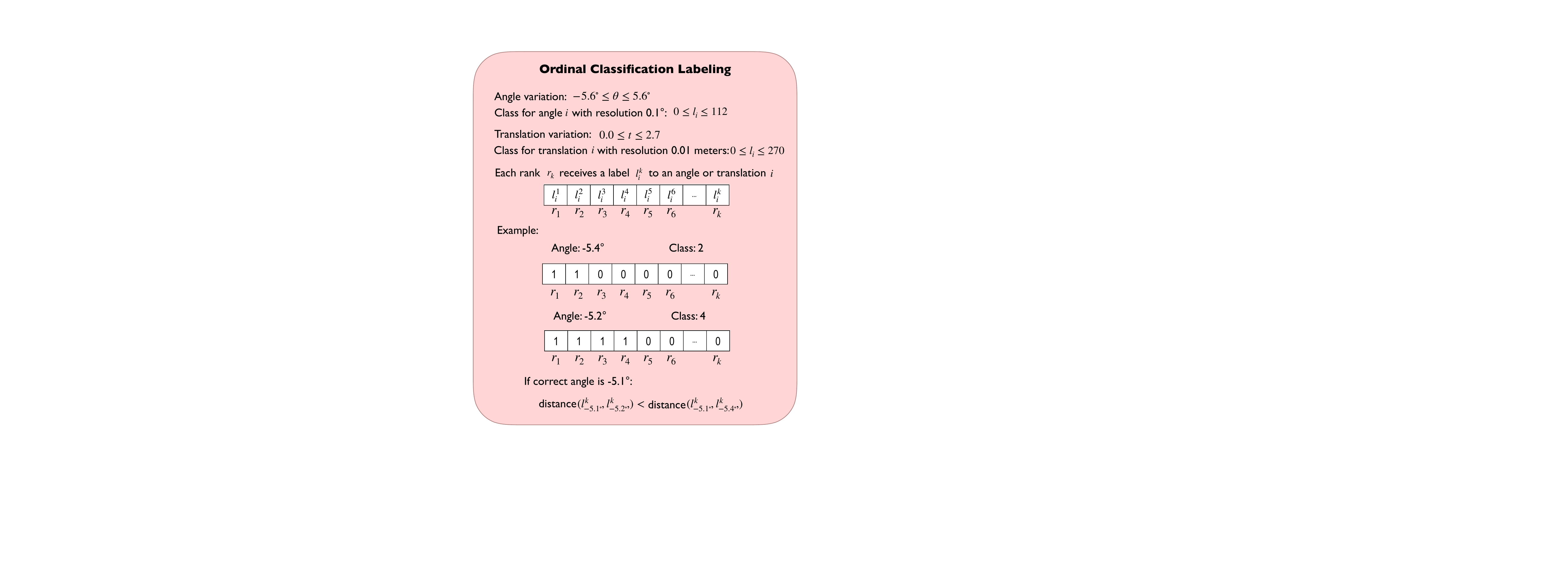}
	\caption{Summary of the ordinal classification labeling with an example for the angle classification. }
	\label{fig:ordinal_classification}
\end{figure}

The rank format allows the network to learn the order of the classes by showing that one value is smaller or larger than other. A summary of the ordinal classification labeling process is presented in \autoref{fig:ordinal_classification}.

To solve the binary classification subproblems we calculate the Binary Cross Entropy between the target and the output for rotation and translation and we sum them as follow:

\begin{equation}
\begin{split}
    &\mathcal{L} = \mathcal{L}_\textit{BCE}({\hat{d}}, {d}) + \beta 
\: \mathcal{L}_\textit{BCE}({{\hat{\theta}},{\theta}})\\
&\text{where } \mathcal{L}_\textit{BCE}(x, y) = - \sum_k{y_k\log{(x_k)} + (1-y_k) \log{(1-x_k)}}
\end{split}
\label{loss_function}
\end{equation}
${{d}}$ and ${\theta}$ are relative ground-truth translation and rotation rank labels, and ${\hat{d}}$ and ${\hat{\theta}}$ their output of the network counterparts. ${\hat{d}}$ and ${\hat{\theta}}$ pass by a Sigmoid function before the loss function for numerical stability. We use the parameter $\beta > 0$ to balance the scale difference between the rotation and translation loss values. 

The network is implemented on the framework PyTorch and the Adam optimizer is applied with learning rate 0.0001. As recommended in \cite{deepvo}, during the pre-training of the CNN-Cam we initialize it with pre-trained FlowNet model weights to reduce the training time. 

%%%%%%%%%%%%%%%%%%%%%%%%%%%%%%%%%%%%%%%%%%%%%%%%%%%%%%%%%%%%%%%%%%%%%%%%%%%%%%
%%%%%%%%%%%%%%%%%%%%%%%%%%%%%%% Results %%%%%%%%%%%%%%%%%%%%%%%%%%%%%%%%%%%%%%

\section{Results}\label{results.sec}

For validation we use the KITTI dataset \cite{kitti}, which provides several sequences in different conditions for outdoor environments. To simulate a 2D laser scanner data, we extracted one $\ang{360}$ layer from the Velodyne data, while for the camera we use the RGB raw images provided by the dataset. We use the 11 sequences from the KITTI odometry dataset which contain ground truth values to train and test the proposed method. Among the 11 sequences, we separate 8 for training and 3 for testing. We use for training the sequences 00, 02, 03, 04, 05, 06, 08 and 09 and for testing the sequences 01, 07 and 10. We chose these three sequences because they are not very long, leaving more data for the training, but they can still be challenging and present the potential of the proposed method. In the previous work \cite{previouswork}, we could not evaluate the sequence 01 because the vehicle is in a highway where the 2D laser scanner could not detect obstacles most of the time, making it impossible for a laser-only network to predict the odometry. For this reason, we add this sequence specially to observe if the fusion could improve the results.

In order to validate our method and compare to the solution DeepVO, we calculated the drift according to the KITTI VO \cite{kitti} evaluation metrics, i.e., averaged Root Mean Square Errrors (RMSEs) of the translation and rotation error for all subsequences (100, 200,..., 800 meters). However, we need to adapt our 2D results to be able to compare to the 3D errors of DeepVO, therefore we create 3D poses from our 2D values by giving 0.0 to the values we do not estimate (lateral and longitudinal angles and translation in the vertical axis).

\begin{table}[]
\centering
\def\arraystretch{1.5}
\begin{tabular}{
>{\columncolor[HTML]{FFFFFF}}c 
>{\columncolor[HTML]{FFFFFF}}c 
>{\columncolor[HTML]{FFFFFF}}c 
>{\columncolor[HTML]{FFFFFF}}c 
>{\columncolor[HTML]{FFFFFF}}c }
\hline
\cellcolor[HTML]{FFFFFF}{\color[HTML]{333333} }                           & \multicolumn{2}{c}{\cellcolor[HTML]{FFFFFF}CNN-Fusion}      & \multicolumn{2}{c}{\cellcolor[HTML]{FFFFFF}DeepVO} \\
\multirow{-2}{*}{\cellcolor[HTML]{FFFFFF}{\color[HTML]{333333} Sequence}} & $t_{rel}$ & $r_{rel}$                                           & $t_{rel}$                  & $r_{rel}$                 \\ \hline
\multicolumn{1}{c|}{\cellcolor[HTML]{FFFFFF}{\color[HTML]{333333} 07}}    & 2.03    & \multicolumn{1}{c|}{\cellcolor[HTML]{FFFFFF}0.85} & 3.91                     & 4.60                    \\
\multicolumn{1}{c|}{\cellcolor[HTML]{FFFFFF}{\color[HTML]{333333} 10}}    & 7.60   & \multicolumn{1}{c|}{\cellcolor[HTML]{FFFFFF}2.80} & 8.11                     & 8.73                    \\ \hline
\multicolumn{1}{c|}{\cellcolor[HTML]{FFFFFF}Mean}                         & 4.41    & \multicolumn{1}{c|}{\cellcolor[HTML]{FFFFFF}1.82} & 6.01                     & 6.66                    \\ \hline
\multicolumn{1}{c|}{\cellcolor[HTML]{FFFFFF}Computation Time (s/frame)}   & \multicolumn{2}{c|}{\cellcolor[HTML]{FFFFFF}0.1}            & \multicolumn{2}{c}{\cellcolor[HTML]{FFFFFF}1.0}   
\end{tabular}
\caption{Average translation (\%) and rotation (\degree$/100m$) RMSE drift on trajectory lenghts of 100 to 800 meters, along with the computation time per frame, for the proposed approach and the DeepVO \cite{deepvo} method.}
\label{table_deepvo}
\end{table}

\autoref{table_deepvo} presents the translation and rotation error score for the testing sequences 07 and 10 and we compare these values between the proposed approach and the method DeepVO, along with their computational times. Even though we transform our results to 3D, it is important to mention that the results are not yet directly comparable, but we can still have an estimation if the order of error is around the same and we can compare the computation time of both methods. For the translation error the comparison is easier to perform since the global translation in the vertical axis is very small and not significant compared to the other axes. On the other hand, for the angles comparison, since most of the time the 3 angles are around 0 degree (the only high rotation rates are on z axis during turns, which represent a very small part of a complete trajectory), all their drifts are in the same order of magnitude and should be relevant for the comparison; For this reason, it is expected that the DeepVO method has a rotation error around three times more than our solution. As a result of these difficulties in the comparison of the two methods, we can only really extract from these results that the errors are around the same order of magnitude but we are still providing a solution that is 10 times faster using no GPU acceleration (2,6 GHz Intel Core i5, Intel Iris 1536 MB), and can be as fast as 0.01s with GPU acceleration (4,0 GHz Intel Core i7, GeForce GTX 1060). This faster processing is mainly due to the fact that we resize the images for smaller sizes and we use only CNNs instead of a RCNN.

\begin{figure}
    \centering
    \begin{subfigure}[b]{0.8\linewidth}
    \centering
    {\includegraphics[width=1.0\linewidth]{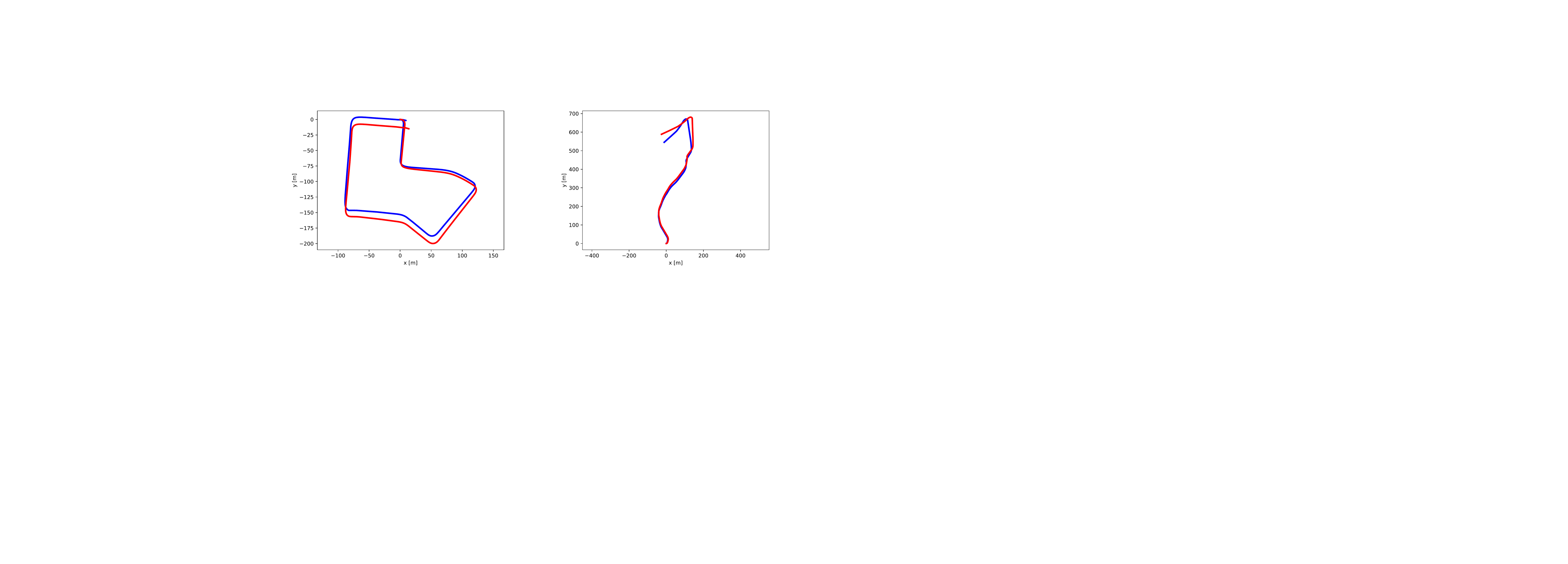} }
    \caption{Sequence 07}
    \end{subfigure}
    \qquad
    \begin{subfigure}[b]{0.8\linewidth}
    \centering
    {\includegraphics[width=1.0\linewidth]{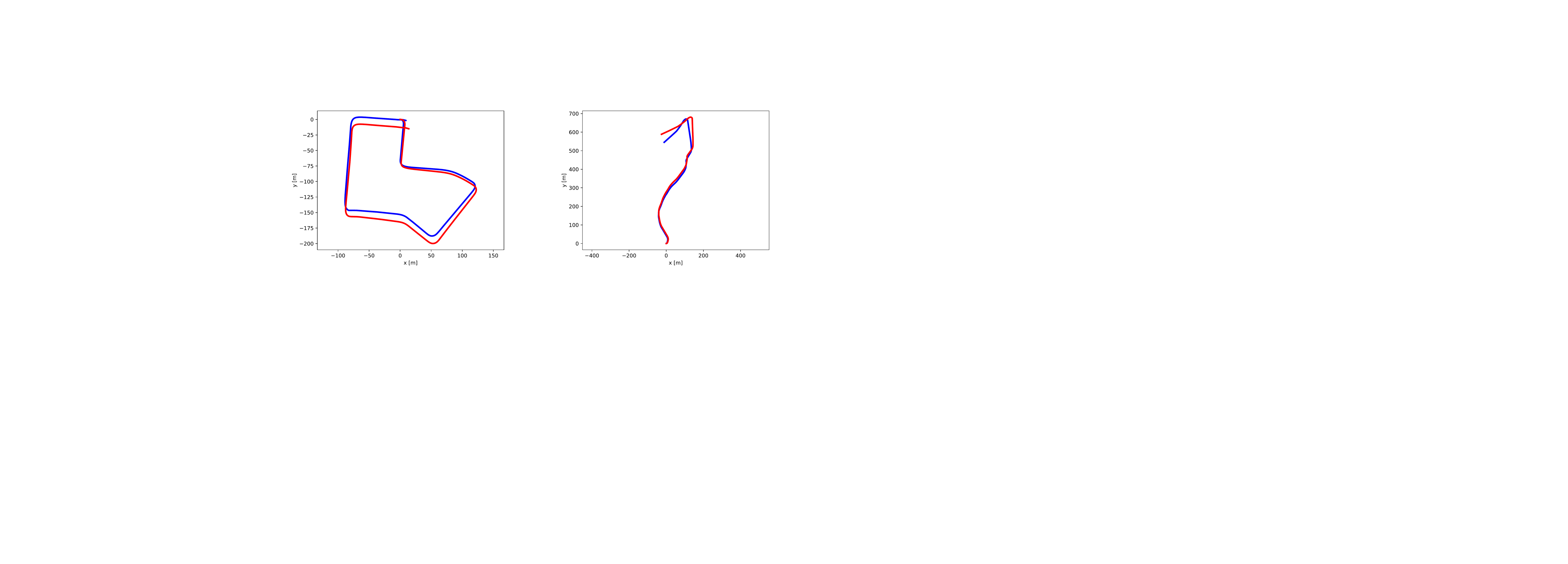} }
    \caption{Sequence 10}
    \end{subfigure}
    \caption{Trajectories of two test sequences (07 and 10) applying the proposed CNN-Fusion. The blue lines represent the ground truth trajectory, while in red the predicted one.}%
    \label{fig:trajectory}%
\end{figure}

\begin{figure*}
\centering
\begin{subfigure}{1.0\textwidth}
  \centering
  \includegraphics[width=1.0\linewidth]{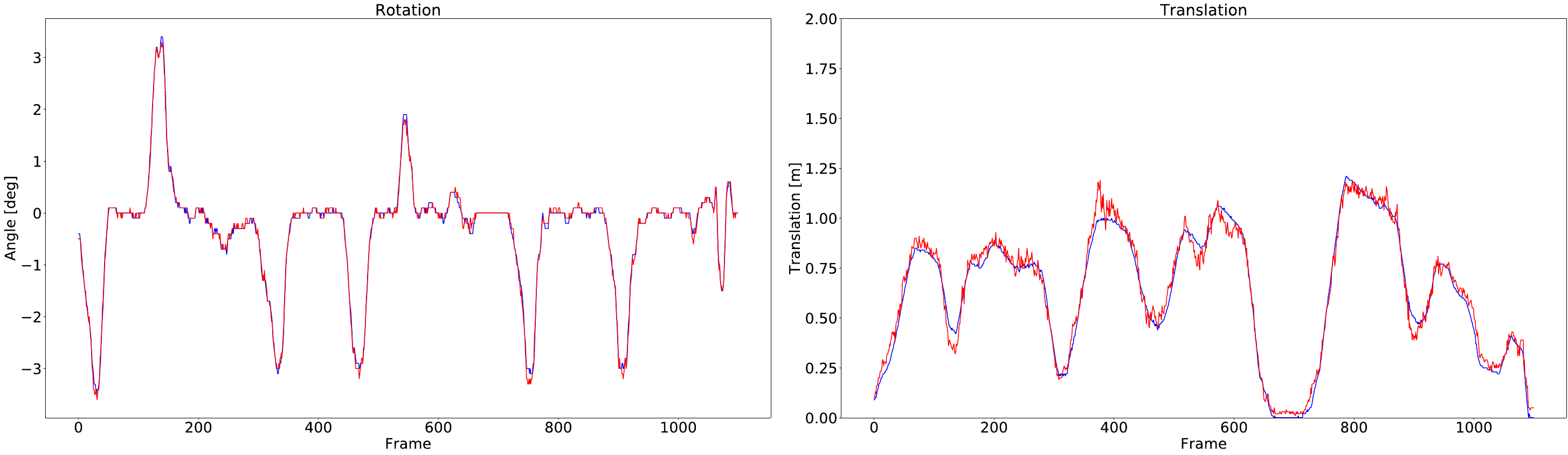}
  \caption{Sequence 07}
  \label{fig:sub1}
\end{subfigure}%
\vspace{0.5cm}
\begin{subfigure}{1.0\textwidth}
  \centering
  \includegraphics[width=1.0\linewidth]{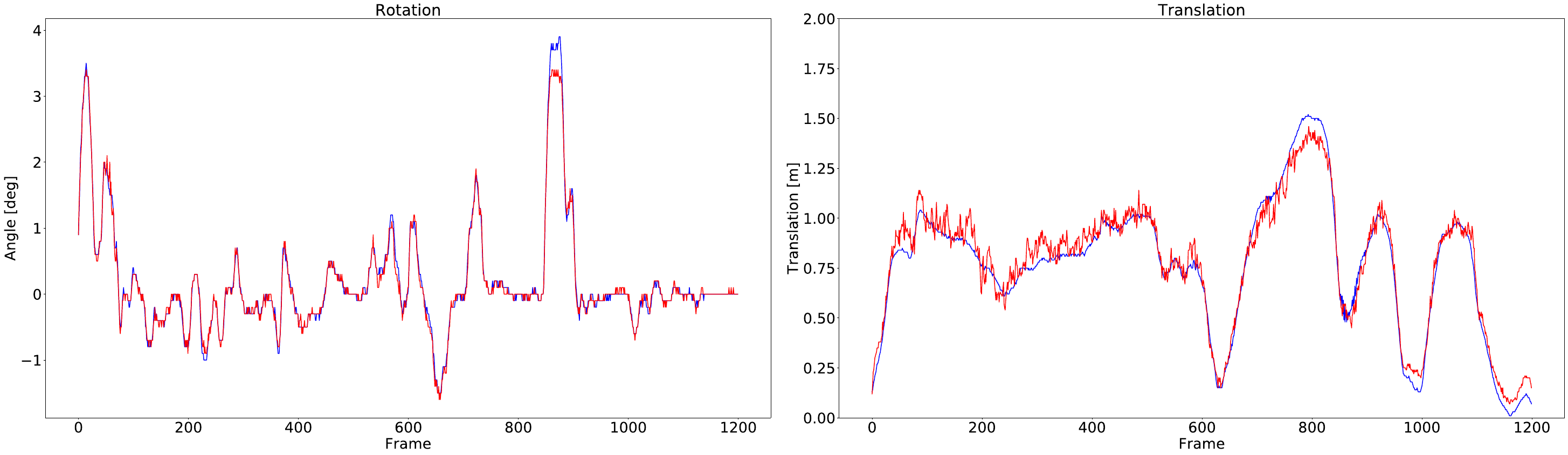}
  \caption{Sequence 10}
  \label{fig:sub2}
\end{subfigure}
\vspace{0.5cm}
\caption{Results of rotation and translation estimation for the testing sequences 07 and 10. The ground truth values are presented in blue and in red the output of the Fusion-CNN network. }
\label{fig:rot_trans}
\end{figure*}

We can observe in Figure \ref{fig:trajectory} that even with the eventual errors in the odometry estimation, we can still obtain a trajectory close to the ground truth. However, since the proposed approach does not perform any sort of loop closure, and does not even use temporal information by recurrent neural networks (as in the DeepVO method), one eventual large error can be accumulated over time, generating a large drift like the one we have by the end of sequence 10.  

\begin{table}[]
\centering
\def\arraystretch{1.5}
\begin{tabular}{ccccc
>{\columncolor[HTML]{EFEFEF}}c 
>{\columncolor[HTML]{EFEFEF}}c }
\hline
                           & \multicolumn{2}{c}{CNN-Cam}            & \multicolumn{2}{c}{CNN-Laser}          & \multicolumn{2}{c}{\cellcolor[HTML]{EFEFEF}CNN-Fusion} \\
\multirow{-2}{*}{Sequence} & $\sigma_r$ & $\sigma_t$                & $\sigma_r$ & $\sigma_t$                & $\sigma_r$                 & $\sigma_t$                \\ \hline
\multicolumn{1}{c|}{01}    & 0.09       & \multicolumn{1}{c|}{0.30} & 0.30       & \multicolumn{1}{c|}{0.33} & 0.06                       & 0.30                      \\
\multicolumn{1}{c|}{07}    & 0.06       & \multicolumn{1}{c|}{0.06} & 0.06       & \multicolumn{1}{c|}{0.03} & 0.04                       & 0.03                      \\
\multicolumn{1}{c|}{10}    & 0.06       & \multicolumn{1}{c|}{0.11} & 0.08       & \multicolumn{1}{c|}{0.04} & 0.05                       & 0.04         
\end{tabular}
\caption{Average rotation absolute error $\sigma_r$ (degrees) and the average translation absolute error $\sigma_t$ (meters) results for the single sensor CNNs, CNN-Cam and CNN-Laser, and the result after CNN-Fusion.}
\label{table_fusion}
\end{table}

For this reason, a better way to understand the accuracy of the proposed approach is presented in \autoref{table_fusion} and in \autoref{fig:rot_trans}. \autoref{table_fusion} shows the average rotation absolute error $\sigma_r$ and the average translation absolute error $\sigma_t$ for the single sensor CNNs and the CNN-Fusion. We can observe that in all the cases the result of the fusion was equal or better than the result of the single sensor network. Specially for the rotation estimation, the fusion of the features was able to increase the accuracy in all of the sequences, which proves that the network was able to learn how to perform the fusion of the two sensors. In the sequence 01, which is the hardest sequence because of the vehicle's velocity and lack of features to detect, it is clear how the laser was not able to estimate the angles because of the few detected points, but after the fusion with the camera the network was able to estimate more accurate angles. However, the translation was still inaccurate because this is the only sequence in the training dataset where the vehicle has a high velocity, therefore there were not other samples for it to learn the translation classes for this case. 

\autoref{fig:rot_trans} presents the odometry estimation (rotation and translation) together with the ground truth for each frame of the sequences 07 and 10. These values present how the network can most of the time estimate accurate odometry, however there are still some difficult cases that can result in inaccurate values. For example, in the sequence 10 we can notice that the highest angle was not properly estimated, this probably happened because there are not a lot of samples for this type of rotation, making it hard for the network to learn this rotation class. We can expect that training this type of network with a larger dataset, with more samples for the challenging cases, could possibly resolve this type of problems and increase the accuracy.

The results show how promising is the fusion between sensors by the use of CNNs, and that the proposed method could be used as a complement to traditional localization methods for intelligent vehicles or any mobile robot. It is also important to mention that we trained the network with a relatively small dataset compared to other deep learning classification tasks, therefore the result could be considerably improved using more sequences for training.

%%%%%%%%%%%%%%%%%%%%%%%%%%%%%%%%%%%%%%%%%%%%%%%%%%%%%%%%%%%%%%%%%%%%%%%%%%%%%%
%%%%%%%%%%%%%%%%%%%%%%%%%%%%%%% Conclusion %%%%%%%%%%%%%%%%%%%%%%%%%%%%%%%%%%%
\section{Conclusion and Future Work}\label{conclusion.sec}

In this paper we presented the first Deep Learning approach for sensor fusion to odometry estimation. We used as input only 2D laser scanner data and camera images to match their features in order to determine the translation and rotation of the vehicle using sequences of CNNs. The proposed network presents that the fusion between the sensors is possible by applying a purely CNN method and we can obtain good accuracy using only low cost sensors. We also introduced a new form of treating the odometry problem in deep learning methods by transforming the regression task into smaller binary classification subproblems that facilitates the training of the network. 

We evaluated the results using the KITTI odometry dataset, making it possible to compare to other approaches. The results showed competitive accuracy, however classic approaches can still provide better results and a better understanding of the quality of their outputs. In spite of that, the proposed approach could be an interesting complement for classic localization estimation methods, since it can be run in real-time and could give relatively accurate values in systems where no wheel encoder data is provided or GPS signal is absent. Moreover, the proposed method presents that the use of Neural Networks is possible to perform the fusion between 2D laser scanners and mono-cameras, and this method could be used for other tasks in robotic systems. In the future work, better results could be obtained by training on larger datasets, specially with samples where the vehicle is moving in high speed and when there are sharp turns, and by exploring the use of temporal information or loop closure methods. 

\addtolength{\textheight}{-12cm}  

%%%%%%%%%%%%%%%%%%%%%%%%%%%%%%%%%%%%%%%%%%%%%%%%%%%%%%%%%%%%%%%%%%%%%%%%%%%%%%

%%%%%%%%%%%%%%%%%%%%%%%%%%%%%%%% Bibliography %%%%%%%%%%%%%%%%%%%%%%%%%%%%%%%%

%%%%%%%%%%%%%%%%%%%%%%%%%%%%%%%%%%%%%%%%%%%%%%%%%%%%%%%%%%%%%%%%%%%%%%%%%%%%%%

\end{document}